\documentclass[a4paper,fleqn]{cas-dc}

\usepackage[numbers]{natbib}
\usepackage{amssymb}
\usepackage{latexsym}
\usepackage{graphicx}
\usepackage{algorithm}
\usepackage{algorithmic}
\usepackage{amsmath} 
\def\tsc#1{\csdef{#1}{\textsc{\lowercase{#1}}\xspace}}
\tsc{WGM}
\tsc{QE}
\tsc{EP}
\tsc{PMS}
\tsc{BEC}
\tsc{DE}

\begin{document}
\let\WriteBookmarks\relax
\def\floatpagepagefraction{1}
\def\textpagefraction{.001}
\shorttitle{Leveraging social media news}
\shortauthors{CV Radhakrishnan et~al.}

\title [mode = title]{Energy Clustering for Unsupervised Person Re-identification}               

\author[1]{Kaiwei Zeng}
\author[1]{Munan Ning}
\author[1]{Yaohua Wang}\cormark[1]
\author[1]{Yang Guo }\cormark[1]
\address[1]{National University of Defense Technology, Changsah, Hunan, China}

\cortext[cor1]{Corresponding author}
\cortext[cor2]{weilantiankong2011@qq.com}

\begin{abstract}
Due to the high cost of data annotation in supervised person re-identification (re-ID) methods, unsupervised methods become more attractive in the real world. Recently, the hierarchical clustering serves as a promising unsupervised method. One key factor of hierarchical clustering is the distance measurement strategy. Ideally, a good distance measurement should consider both inter-cluster and intra-cluster distance of all samples. To solve this problem, we propose to use the energy distance to measure inter-cluster distance in hierarchical clustering (E-cluster), and use the sum of squares of deviations (SSD) as a regularization term to measure intra-cluster distance for further performance promotion. We evaluate our method on Market-1501 and DukeMTMC-reID. Extensive experiments show that E-cluster obtains significant improvements over the state-of-the-arts fully unsupervised methods.
\end{abstract}

\begin{keywords}
person re-identification; fully unsupervised method; hierarchical clustering; energy distance
\end{keywords}

\maketitle

\section{Introduction}
Person re-identification (re-ID) is a task about whether a person reappears in another camera after captured by one, which is widely used in the field of security. With the development of deep learing, supervised methods \citep{sun2018beyond,wang2018learning} achieve good performance in re-ID. But supervised methods need lots of manual labels, which is very expensive in real life.To mitigate above problems, people focus on unsupervised domain adaptation (UDA) which only needs labeled source data \citep{wei2018person,zhong2018generalizing,zhong2019invariance,peng2016unsupervised,tian2019person,xu2019learning}. But in fact, it is not strictly fully unsupervised methods, which should further remove the need of labels about the source dataset.

\begin{figure}
	\centering
		\includegraphics[scale=.3]{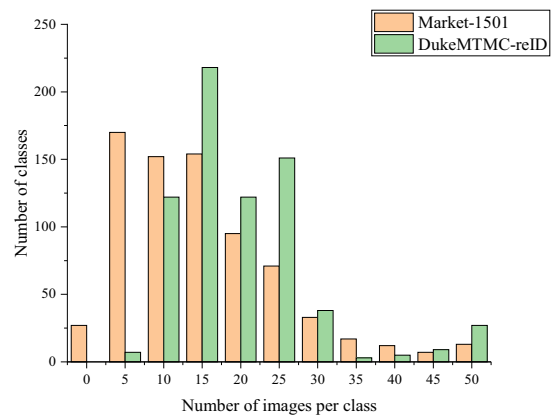}
	\caption{The distribution of Market-1501 and DukeMTMC-reID are uneven and different from each other.}
	\label{fig1}
\end{figure}

For UDA, Fan et al. propose PUL \citep{fan2018unsupervised}, they use K-means to cluster different samples in each iteration, regard clusters numbers $1,\cdots,k$ as pseudo labels, and fine-tune model with pseudo labels. However, $k$ value needs to be determined in advance and the result of the K-means is sensitive to the $k$ value. To improve the performance of PUL, Lin et al. propose a fully unsupervised method BUC \citep{lin2019bottom}. They use hierarchical clustering to merge a fixed number of clusters and fine-tune the model in each iteration. BUC uses the minimum as the distance measurement. It only calculates the nearest pairwise distance of one pair of samples between two clusters, but ignores features of other samples. Besides, BUC use samples number in each cluster as the regularization term to avoid poor clusters. However, as shown in Figure \ref{fig1}, the distribution of samples in datasets is not uniform, so it's also not a good choice to simply use samples number as the regularization term. We should allow uneven clusters.

In general, a good distance measurement should consider both inter-cluster and intra-cluster distance of all samples. Otherwise, some important information will be ignored and it will result in poor clustering results. Hierarchical clustering uses euclidean distance in calculation. According to the conclusion in \citep{wang2017normface,ding2019towards}, it is easy to form elongate clusters which result in poor performance when using the minimum distance. 

\begin{figure*}
	\centering
		\includegraphics[scale=.6]{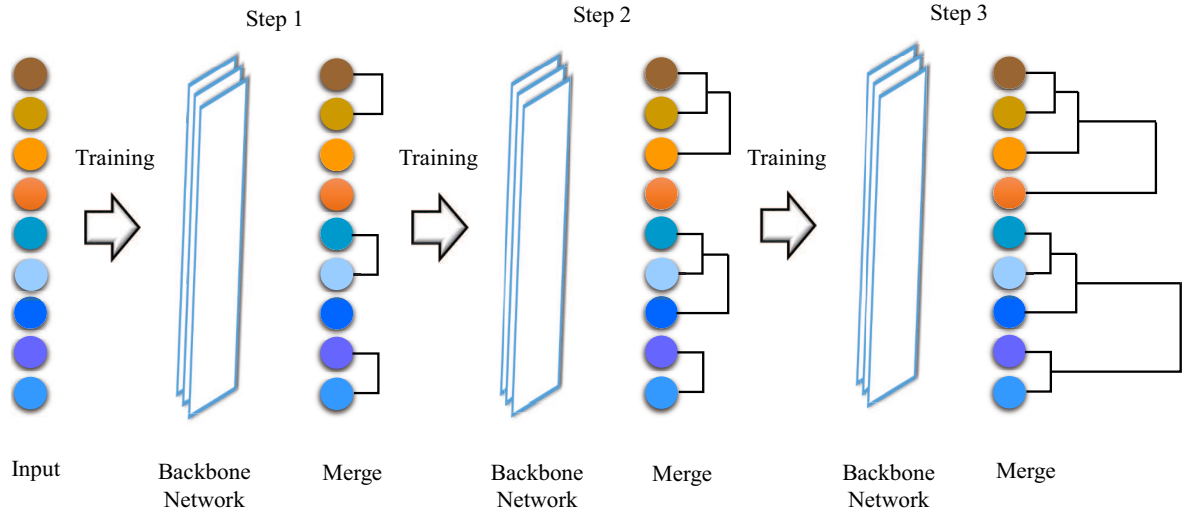}
	\caption{Hierarchical clustering. Each circle represents a sample, we use ResNet-50 as backbone network to extract features. According to cluster similarities, we will merge samples in each step. Finally, we update pseudo labels so that samples in the same cluster have the same pseudo label.}
	\label{fig2}
\end{figure*}

Based on BUC, Ding et al. propose DBC \citep{ding2019towards}. They use the unweighted average linkage clustering
(unweighted pairgroup method with arithmetic means, UPGMA) as the distance measurement to solve these problems. However, DBC does not consider the special situation that when samples in ${C}_{a}$,${C}_{b}$ are the same (${C}_{i}$ means clusters in sample space), distance between ${C}_{a}$,${C}_{b}$ should be zero. More importantly, this distance will increase with sample number and the dispersion degree in clusters. When both ${C}_{b}$,${C}_{c}$ are close to ${C}_{a}$ in high-dimensional space, this distance is enough to influence the selection of merging, finally influence model performance. Besides, for regularization term, BUC only consider samples number in each cluster, DBC only consider the dispersion degree in each cluster. However, a better method should combine them together.

In order to solve these problems, we propose to use energy distance \citep{szekely2013energy,szekely2003statistics} as the distance measurement to evaluate inter-cluster distance and use the sum of squares of deviations (SSD) to measure intra-cluster distance. Energy distance not only calculate pairwise distance of all samples in clusters, but also has a good property: if and only if ${C}_{a}$,${C}_{b}$ are the same, the distance between ${C}_{a}$,${C}_{b}$ is zero. This property can promote more compact clustering. Moreover, we introduce the SSD as a regularization term. SSD considers both the number of samples and the dispersion degree in each cluster which can better evaluate intra-cluster distance and further improve model performance. 

To summarize, the major contributions of this paper are:
\begin{quote}
\begin{itemize}
\item We measure the distance between clusters with energy distance, which can promote more compact clustering.
\item We use SSD as the regularization term, it gives priority to merge single sample clusters and also allows uneven clustering, which can further improve model performance.
\item Experiment results show that our method achieves the state-of-the-arts on Market-1501 and DukeMTMC-reID in fully unsupervised methods.
\end{itemize}
\end{quote}

\begin{figure*}
	\centering
		\includegraphics[scale=.45]{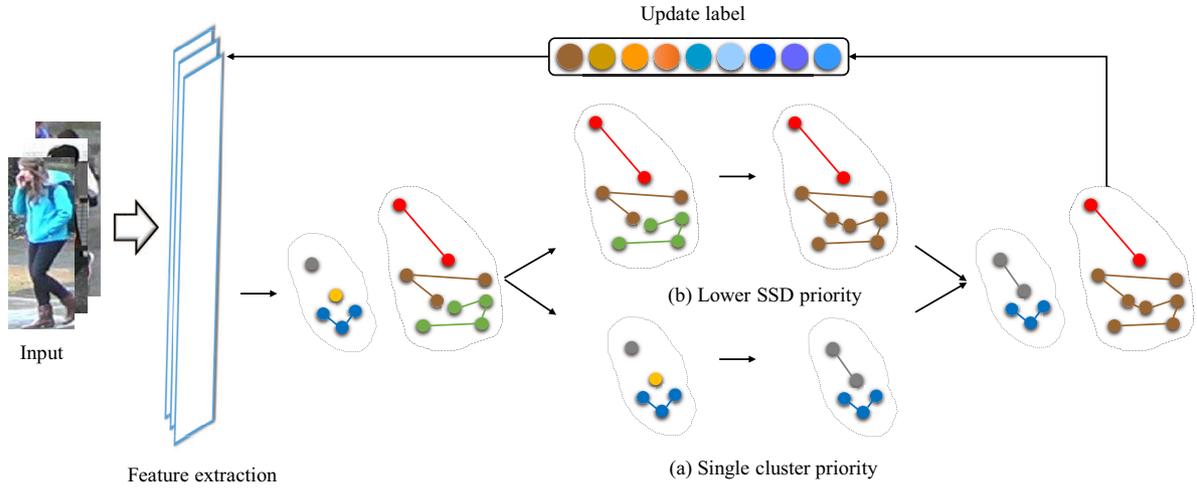}
	\caption{(a) for the single-sample cluster, it will be merged preferentially because its SSD is zero. So the yellow and the gray sample will be merged. (b) The SSD measures the dispersion of samples in clusters and it is not linearly dependent on the samples number. As a result, it does not exclude the formation of super clusters. As shown in the figure, the brown cluster will merge the green cluster, instead of the red cluster, because it has a lower SSD.}
	\label{fig3}
\end{figure*}

\section{Related Work}
\subsection{Unsupervised Domain Adaptation}
Some UDA methods focus on differences of camera styles in different datasets. They use GAN to generate new images \citep{wang2018transferable,qian2018pose,zheng2019joint}. On the one hand, it can generate new images with labels and decrease differences caused by camera styles. On the other hand, it can also expand training datasets. Deng et al. propose SPGAN \citep{deng2018image} based on CycleGAN \citep{zhu2017unpaired}. The core idea of SPGAN is that identities of images can keep the same before and after transferring. They directly transfer the style of images in the source domain to the target domain. Finally, they use new generated images for training.

Others use auxiliary information to generate pseudo labels for images and conduct training with these labels. Zhong et al. propose ECN \citep{zhong2019invariance} based on exemplar-invariance \citep{wu2018unsupervised,xiao2017joint},  camera-invariance \citep{zhong2018generalizing} and neighborhood-invariance \citep{chen2018deep}. They set triplet loss about these invariances and use them to increase the distance between different samples and reduce the distance between similar samples. They use exemplar memory module \citep{santoro2016meta,vinyals2016matching} to set pseudo labels for samples, and use these pseudo labels to optimize the model in each iteration. 

\subsection{Clustering-guided Re-ID}
\subsubsection{Clustering-guided Domain Adaptation}
Fan et al. propose PUL \citep{fan2018unsupervised}. They extract features through convolutional neural network (CNN) and divide samples into cluster $C_{i}$ according to the distance between samples and clustering centroids. Finally, they regard $i$ as the pseudo label of samples. At the beginning, when the model is poor, PUL learns only from a small amount of reliable samples which are close to the cluster centroid to avoid falling into local optimal. As the model becomes more and more better, more samples will be selected in training. Finally, the model repeats CNN fine-tuning and K-means clustering until convergence. Howerver, PUL is not stable because K-means is sensitive to $k$ value and $k$ is difficult to determine.

\subsubsection{Fully Unsupervised Methods}
Lin et al. propose BUC \citep{lin2019bottom}. Specifically, (1) they use CNN to extract features and generate a sample space, (2) they merge a fixed percent of similar clusters in each iteration according to the distance between clusters and update pseudo labels according to clustering results, (3) they fine-tune the model according to pseudo labels. The distance measurement used in hierarchical clustering is the most important because it decides clustering results and pseudo labels, further decides the model performance. BUC compares the minimum distance, maximum distance, and average distance, finally chooses the minimum distance to measure inter-cluster distance and takes samples number $n_{i}$ in each cluster as the regularization term to avoid poor and super clusters.

However, the minimum distance only selects one pair of the nearest samples in two clusters without considering other samples in clusters. Besides, samples number ignores the dispersion degree in clusters. Based on BUC, Ding et al. propose DBC \citep{ding2019towards}. They use UPGMA to measure inter-cluster distance, use pairwise distances of all samples in each cluster to measure intra-cluster distance. Finally, they get significant improvement on BUC. But UPGMA cannot satisfy the distance between ${C}_{a}$,${C}_{b}$ should be zero when samples in ${C}_{a}$,${C}_{b}$ are the same. Besides, it only considers the dispersion degree in intra-cluster distance which is not enough. Our E-cluster resolves these problems and get better performance.

\section{Model Architecture}
\subsection{Hierarchical Clustering}
A given training set $X =\left\{x_{1},x_{2},\cdots,x_{N}\right\}$, we have manual annotation $Y =\left\{y_{1},y_{2},\cdots,y_{n}\right\}$ in supervised methods. Therefore, we can directly take these labels as supervision to conduct model training with cross entropy. 
In general, we usually use softmax function to predict image identities:
\begin{equation}
\label{equ:E1}
P\left(y_{i} | x\right)=\frac{\exp \left(w_{i}^{T} x\right)}{\sum_{j=1}^{n} \exp \left(w_{j}^{T} x\right)}
\end{equation}
where $y_{i}$ is the predicted label, $P\left(y_{i} | x\right)$ is the predicted probability of $x$ belonging to $y_{i}$, $w$ is the weight of model. We can learn $y=f\left(w,x\right)$ directly from CNN in supervised methods . 

But in unsupervised methods, there is no any manual annotation like $y_{i}$, we need to learn a feature embedding function $\phi\left(\theta ; x_{i}\right)$, where $\theta$ is the weight of CNN. For the query set $\left\{x_{i}^{q}\right\}_{i=1}^{N_{q}}$ and the gallery set $\left\{x_{i}^{g}\right\}_{i=1}^{N_{g}}$, we need to compare the similarity of images to determine the image identity according to the distance between features: $d\left(x_{i}^{q}, x_{i}^{g}\right)=\left\|\phi\left( \theta;x_{i}^{q} \right)-
\phi\left(\theta; x_{i}^{g}\right)\right\|$, Then we can generate pseudo labels for images and use these labels to conduct training. Our network structure is shown in Figure \ref{fig2}, which mainly includes two parts: fine-tuning and clustering. We repeat them until converge. We define the probability that image $x$ belongs to the $i$-$th$ cluster as:
\begin{equation}
\label{equ:E2}
p(i | x, \boldsymbol{V})=\frac{\exp \left(\boldsymbol{V}_{i}^{\mathrm{T}} \boldsymbol{v} / \tau\right)}{\sum_{j=1}^{n} \exp \left(\boldsymbol{V}_{j}^{\mathrm{T}} \boldsymbol{v} / \tau\right)}
\end{equation}
where $\boldsymbol{V}$ is the matrix that contains all features, $\boldsymbol{V}_{i}$ is the $i$-$th$ cluster, $n$ is the cluster number in the current iteration, $\boldsymbol{v}=\frac{\phi(\theta ; x)}{\|\phi(\theta ; x)\|}$ is the $L_{2}$ normalized feature. $\boldsymbol{V}_{j} \cdot \boldsymbol{v}$ means the cosine distance between $x$ and cluster $C_{i}$, $\tau$ is a temperature parameter \citep{hinton2015distilling} that controls the softness of probability distribution. According to \citep{xiao2017joint}, we set $\tau= 0.1$.

\subsection{Energy Distance}
In hierarchical clustering, how to measure the distance between clusters directly determines the performance of the model. Different from other methods, energy distance is the distance between statistical observations in metric spaces. This concept derives from Newton's gravitational potential energy. 

Energy distance views the sample as an object subject to statistical potential energy. According to theories in \citep{szekely2012contests,szekely2013energy}, we measure the distance by:
\begin{equation}
\label{equ:E3}
2 \int_{-\infty}^{\infty}(F(x)-G(x))^{2} d x
\end{equation}
where $F(\cdot )$, $G(\cdot )$ are cumulative distribution functions (cdf) of a random variable. For $X, Y$ are independent random variables with cdf $F$ and $G$:
\begin{align}
&2 \int_{-\infty}^{\infty}(F(x)-G(x))^{2} d x \nonumber\\
=\quad &2 E|X-Y|-E\left|X-X^{\prime}\right|-E\left|Y-Y^{\prime}\right|
\end{align}
where $X^{'}$ is an independent and identically distributed ($iid$) copy of $X$, $Y^{'}$ is an $iid$ copy of $Y$, $E(\cdot )$ is the mathematical expectation of samples. When it is extended to the rotation invariant high dimensional space, the energy distance is:
\begin{equation}
\label{equ:E5}
\mathcal{E}(X, Y)=2 E|X-Y|_{d}-E\left|X-X^{\prime}\right|_{d}-E\left|Y-Y^{\prime}\right|_{d}
\end{equation}
where $X, Y\in R^{d}$ are two $iid$ sample space, $d$ is the dimension of the sample space. Sz{\'e}kely et al. prove that Eq.(\ref{equ:E5}) is nonnegative and equals zero if and only if $X$ and $Y$ are $iid$ in \citep{szekely2005new}. When it is extended to multi-sample energy distance, that's what we used in E-cluster:
\begin{align}
\mathcal{E}(X, Y)=&\frac{2}{n m} \sum_{i, j=1}^{n, m}\left\|x_{i}-y_{j}\right\|_{2}  \\ &-\frac{1}{n^{2}}\sum_{i,j=1}^{n}\left\|x_{i}-x_{j}\right\|_{2}-\frac{1}{m^{2}} \sum_{i, j=1}^{m}\left\|y_{i}-y_{j}\right\|_{2} \nonumber
\end{align}
where $X=\left\{x_{1}, x_{2}, \cdots, x_{n}\right\}$, $Y=\left\{y_{1},y_{2},\cdots,y_{m}\right\}$ and $n$,$m$ are samples number in $X,Y$. $\left\|\cdot\right\|_{2}$ means the euclidean distance between samples. 

The ability of energy distance to separate and identify clusters with equal or near equal centers is an important practical advantage over geometric clustering center methods, such as centroid, minimum, maximum and Ward's minimum variance methods. According to \citep{szekely2005hierarchical}, energy clustering can effectively recover the underlying hierarchical structure under different scenarios, including high dimensional data and data with different scale attributes, while maintaining the advantages of separating spherical clustering. So compared with other distance measurement, it can promote more compact clustering and get better results.

\begin{algorithm}[ht]
\caption{$E$-cluster Algorithm}
\begin{algorithmic}[1]
\REQUIRE ~~\\
Input $X=\left\{x_{i}\right\}_{i=1}^{N}$; \\
Merging percent $m \in(0,1)$; \\
Hyperparameter $\lambda$; \\
Initial model $\phi\left(\cdot ; {\theta}_{0}\right)$.
\ENSURE ~~\\
Best model $\phi\left(\cdot ; {\theta}\right)$.
\STATE Initialize: \\
\qquad pseudo labels $Y=\left\{y_{i}=i\right\}_{i=1}^{N}$, \\
\qquad cluster number $n=N$, \\
\qquad merging number $m=n*mp$. \\
\WHILE {$n>m$}
\STATE Train model with $X$,$Y$;
\STATE Extract feature and update $\boldsymbol{V}$;
\STATE Calculate energy distance and SSD between clusters in $\boldsymbol{V}$ according to Eq.(\ref{equ:E8});
\STATE Select clusters to merge: \\
\qquad $n=n-m$;
\STATE Update $Y$ with new pseudo labels:\\
\qquad $Y=\left\{y_{i}=j, \quad \forall x_{i} \in {C}_{j}\right\}_{i=1}^{N}$;
\STATE Evaluate model performance.
\IF{$mAP_{i}>mAP_{best}$}
\STATE $mAP_{best}=mAP_{i}$;
\STATE Best model = $\phi(x ;\theta_{i})$;
\ENDIF
\ENDWHILE 
\label{code:recentEnd}
\end{algorithmic}
\end{algorithm}

\subsection{Regularization Term}
The sum of squares of deviations (SSD) is usually used to measure the sum deviation between each sample and the mean. In other words, it equals to variance$*n$. For sample space $X=\left\{x_{1}, x_{2}, \cdots, x_{n}\right\}$, the mean is $\overline{X}=\frac{1}{n} \sum_{i=1}^{n}x_{i}$, SSD is:
\begin{equation}
\label{equ:E7}
SSD(X)=\sum_{i=1}^{n}(x_{i}-\overline{X})^{2}
\end{equation}
At the beginning, SSD in the single-sample cluster is zero, so we can merge the single-sample cluster first. Besides, it doesn't exclude the form of super clusters because SSD considers both samples number and the dispersion of samples in clusters. In order to balance the SSD and energy distance, we introduce a trade-off parameter $\lambda$, and finally our distance measurement is defined as:
\begin{equation}
\label{equ:E8}
D(X, Y)=\mathcal{E}(X, Y)+\lambda SSD(X)
\end{equation}
where $D(X, Y)$ means distance between cluster $X$ and cluster $Y$. As shown in Figure.\ref{fig3}, our E-cluster can promote lower SSD and single cluster priority during the merging, which can further improve the model performance.

\begin{table*}[width=1.0\textwidth,cols=4,pos=h]
\centering
\caption{Comparison of E-cluster with BUC and DBC. We use bold type to highlite the best performance in all tables.}
\begin{tabular}{lccccc}
\hline
\multicolumn{1}{|l|}{\begin{tabular}[c]{@{}l@{}} \\
Method\end{tabular}} & \multicolumn{2}{c|}{Market-1501} & \multicolumn{2}{c|}{DukeMTMC-reID}  \\ \cline{2-5} 
\multicolumn{1}{|l|}{} & \multicolumn{1}{c|}{rank-1} & \multicolumn{1}{l|}{mAP} & \multicolumn{1}{c|}{rank-1} & \multicolumn{1}{c|}{mAP}\\ \hline
\multicolumn{1}{|l|}{BUC without regularizer \citep{lin2019bottom}} & \multicolumn{1}{c}{62.9} & \multicolumn{1}{c|}{33.8} & \multicolumn{1}{c}{41.3} & \multicolumn{1}{c|}{22.5} \\
\multicolumn{1}{|l|}{BUC with regularizer \citep{lin2019bottom}} & \multicolumn{1}{c}{66.2} & \multicolumn{1}{c|}{38.3} & \multicolumn{1}{c}{47.4} & \multicolumn{1}{c|}{27.5} \\ 
\multicolumn{1}{|l|}{DBC without regularizer \citep{ding2019towards}} & \multicolumn{1}{c}{66.2} & \multicolumn{1}{c|}{38.7} & \multicolumn{1}{c}{48.2} & \multicolumn{1}{c|}{27.5}  \\ 
\multicolumn{1}{|l|}{DBC with regularizer \citep{ding2019towards}} & \multicolumn{1}{c}{69.2} & \multicolumn{1}{c|}{41.3} & \multicolumn{1}{c}{51.5} & \multicolumn{1}{c|}{30.0}  \\ 
\multicolumn{1}{|l|}{E-cluster without regularizer} & \multicolumn{1}{c}{68.5} & \multicolumn{1}{c|}{40.7} & \multicolumn{1}{c}{50.7} & \multicolumn{1}{c|}{28.6} \\
\multicolumn{1}{|l|}{E-cluster with regularizer} & \multicolumn{1}{c}{\textbf{70.2}} & \multicolumn{1}{c|}{\textbf{43.0}} & \multicolumn{1}{c}{\textbf{53.2}} & \multicolumn{1}{c|}{\textbf{31.1}} \\ \hline
\end{tabular}
\label{table1}
\end{table*}

\begin{table*}[width=1.0\textwidth,cols=4,pos=h]
\centering
\caption{We compare E-cluster with recent methods on Market-1501 and DukeMTMC-reID, the label column lists the type of supervision used by the method. "Transfer" means it uses a labeled source dataset, "OneEx" means only one image in per identity is labeled, "None" means no any manually labeled data are used. "*" means results are reproduced by \citep{lin2019bottom}. }
\begin{tabular}{lllcccccccc}
\hline
\multicolumn{1}{|l|}{\begin{tabular}[c]{@{}l@{}} \\ Methods\end{tabular}}  & \multicolumn{1}{l|}{\begin{tabular}[c]{@{}l@{}} \\ Labels\end{tabular}} & \multicolumn{4}{c|}{Market-1501} & \multicolumn{4}{c|}{DukeMTMC-reID} \\ \cline{3-10} 
\multicolumn{1}{|l|}{} & \multicolumn{1}{l|}{} & \multicolumn{1}{c|}{rank-1} & \multicolumn{1}{c|}{rank-5} & \multicolumn{1}{c|}{rank-10} & \multicolumn{1}{c|}{mAP} & \multicolumn{1}{c|}{rank-1} & \multicolumn{1}{c|}{rank-5} & \multicolumn{1}{c|}{rank-10} & \multicolumn{1}{c|}{mAP} \\ \hline
\multicolumn{1}{|l|}{BOW\citep{zheng2015scalable}'15} & \multicolumn{1}{l|}{None} & \multicolumn{1}{c}{35.8} & \multicolumn{1}{c}{52.4} & \multicolumn{1}{c}{60.3} & \multicolumn{1}{c|}{14.8} & \multicolumn{1}{c}{17.1} & \multicolumn{1}{c}{28.8} & \multicolumn{1}{c}{34.9} & \multicolumn{1}{c|}{8.3} \\ 
\multicolumn{1}{|l|}{OIM*\citep{xiao2017joint}'17} &\multicolumn{1}{l|}{None} & \multicolumn{1}{c}{38.0} & \multicolumn{1}{c}{58.0} & \multicolumn{1}{c}{66.3} & \multicolumn{1}{c|}{14.0} & \multicolumn{1}{c}{24.5} & \multicolumn{1}{c}{38.8} & \multicolumn{1}{c}{46.0} & \multicolumn{1}{c|}{11.3} \\ 
\multicolumn{1}{|l|}{UMDL\citep{peng2016unsupervised}'16} & \multicolumn{1}{l|}{Transfer} & \multicolumn{1}{c}{34.5} & \multicolumn{1}{c}{52.6} & \multicolumn{1}{c}{59.6} & \multicolumn{1}{c|}{12.4} & \multicolumn{1}{c}{18.5} & \multicolumn{1}{c}{31.4} & \multicolumn{1}{c}{37.6} & \multicolumn{1}{c|}{7.3} \\ 
\multicolumn{1}{|l|}{PUL\citep{fan2018unsupervised}'18} &  \multicolumn{1}{l|}{Transfer} & \multicolumn{1}{c}{44.7} & \multicolumn{1}{c}{59.1} & \multicolumn{1}{c}{65.6} & \multicolumn{1}{c|}{20.1} & \multicolumn{1}{c}{30.4} & \multicolumn{1}{c}{46.4} & \multicolumn{1}{c}{50.7} & \multicolumn{1}{c|}{16.4} \\ 
\multicolumn{1}{|l|}{EUG\citep{wu2019progressive}'19} & \multicolumn{1}{l|}{OneEx} & \multicolumn{1}{c}{49.8} & \multicolumn{1}{c}{66.4} & \multicolumn{1}{c}{72.7} & \multicolumn{1}{c|}{22.5} & \multicolumn{1}{c}{45.2} & \multicolumn{1}{c}{59.2} & \multicolumn{1}{c}{63.4} & \multicolumn{1}{c|}{24.5} \\ 
\multicolumn{1}{|l|}{SPGAN\citep{deng2018image}'18} &\multicolumn{1}{l|}{Transfer} & \multicolumn{1}{c}{58.1} & \multicolumn{1}{c}{76.0} & \multicolumn{1}{c}{82.7} & \multicolumn{1}{c|}{26.7} & \multicolumn{1}{c}{49.6} & \multicolumn{1}{c}{62.6} & \multicolumn{1}{c}{68.5} & \multicolumn{1}{c|}{26.4} \\ 
\multicolumn{1}{|l|}{TJ-AIDL\citep{wang2018transferable}'18} & \multicolumn{1}{l|}{Transfer} & \multicolumn{1}{c}{58.2} & \multicolumn{1}{c}{-} & \multicolumn{1}{c}{-} & \multicolumn{1}{c|}{26.5} & \multicolumn{1}{c}{44.3} & \multicolumn{1}{c}{-} & \multicolumn{1}{c}{-} & \multicolumn{1}{c|}{23.0} \\ 
\multicolumn{1}{|l|}{HHL\citep{zhong2018generalizing}'18} & \multicolumn{1}{l|}{Transfer} & \multicolumn{1}{c}{62.2} & \multicolumn{1}{c}{78.8} & \multicolumn{1}{c}{84.0} & \multicolumn{1}{c|}{31.4} & \multicolumn{1}{c}{46.9} & \multicolumn{1}{c}{61.0} & \multicolumn{1}{c}{66.7} & \multicolumn{1}{c|}{27.2} \\ 
\multicolumn{1}{|l|}{BUC\citep{lin2019bottom}'19}  & \multicolumn{1}{l|}{None} & \multicolumn{1}{c}{66.2} & \multicolumn{1}{c}{79.6} & \multicolumn{1}{c}{84.5} & \multicolumn{1}{c|}{38.3} & \multicolumn{1}{c}{47.4} & \multicolumn{1}{c}{62.6} & \multicolumn{1}{c}{68.4} & \multicolumn{1}{c|}{27.5} \\ 
\multicolumn{1}{|l|}{DBC\citep{ding2019towards}'19}  & \multicolumn{1}{l|}{None} & \multicolumn{1}{c}{69.2} & \multicolumn{1}{c}{83.0} & \multicolumn{1}{c}{87.8} & \multicolumn{1}{c|}{41.3} & \multicolumn{1}{c}{51.5} & \multicolumn{1}{c}{64.6} & \multicolumn{1}{c}{70.1} & \multicolumn{1}{c|}{30.0} \\ 
\multicolumn{1}{|l|}{E-cluster} & \multicolumn{1}{l|}{None} & \multicolumn{1}{c}{\textbf{70.2}} & \multicolumn{1}{c}{\textbf{84.0}} & \multicolumn{1}{c}{\textbf{88.5}} & \multicolumn{1}{c|}{\textbf{43.0}} & \multicolumn{1}{c}{\textbf{53.2}} & \multicolumn{1}{c}{\textbf{66.2}} & \multicolumn{1}{c}{\textbf{72.0}} & \multicolumn{1}{c|}{\textbf{31.1}} \\ \hline
\end{tabular}
\label{table2}
\end{table*}

\subsection{Update and Merge}
As shown in the algorithm, we regard $N$ samples as $N$ different identities at the beginning. We extract features and measure the distance of each cluster according to energy distance and SSD. Then we merge a fixed percent clusters in each step to generate new clusters and update pseudo labels. Finally, we evaluate model performance in each step. We set two hyperparameters, one is $\lambda$, which is used to balance energy distance and SSD, the other is $mp$, which is used to control merging speed. Besides, $n$ represents the current cluster number, $m=N*mp$ represents the number of clusters merged in each step. That is, $m$ pairs of nearest clusters will be merged in each step. We iterate the model until the model performance no longer improves.

\section{Experiment Results}
\subsection{Datasets}
\subsubsection{Market-1501} Market-1501 \citep{zheng2015scalable} is consist of 1,501 identities observed under 6 camera viewpoints. Each pedestrian is captured by at least two cameras and may has multiple images in a single camera. The training set contains 751 identities about 12,936 images, and the test set contains 750 identities about 19,732 images.

\subsubsection{DukeMTMC-reID}
DukeMTMC \citep{ristani2016performance} contains an 85 minutes high-resolution video from eight different cameras. DukeMTMC-reID \citep{zheng2017unlabeled} is a subset of DukeMTMC. It contains 16,522 images for training, 2,228 images for query, and 17,661 images for gallery.

\subsection{Experiment Setting}
\subsubsection{Evaluation} 
We use the mean average precision (mAP) and the rank-$k$ accuracy to evaluate model performance. The mAP is calculated according to the precision-recall curve, reflecting the overall accuracy and recall rate. Rank-$k$ emphasizes the accuracy of retrieval, it means the query picture has the match in the top-$k$ list.

\subsubsection{Experiment Details} 
In the experiment, we use Resnet-50 \citep{he2016deep} pre-trained on ImageNet as the backbone network. We remove the last classification layer and add a FC layers behind it as the embedding layer. We keep embedding layer still 2048-d. During the training, we set the number of training epochs in the first stage to 20 and in the following stage to 2 for fine-tuning. We set dropout rate to 0.5 and model is trained using stochastic gradient descent (SGD) with the momentum of 0.9. The learning rate is initialized to 0.1 and decreased to 0.01 after 15 epochs. For clustering, we set the merging percent $m$ to 0.05, $\lambda$ to 0.9. We set batch size to 16 in training and change it to 64 in evaluation. 

\begin{figure*}
	\centering
		\includegraphics[scale=.55]{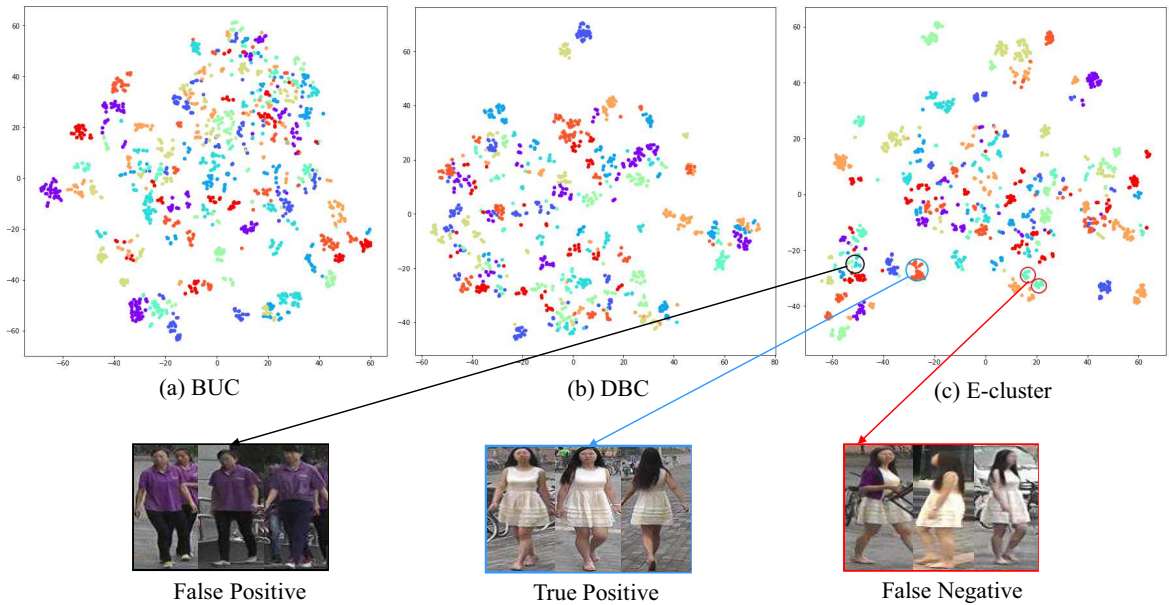}
	\caption{T-SNE visualization of Market-1501 about BUC, DBC, and E-cluster. We choose a subset of 100
identities for visualization. Different color means different real labels. Samples close to each other means they have the same pseudo label. ”True Positive” means model generate correct pseudo labels. ”False Positive” means different people but looks similar. ”False Negative” means the same people but looks different. Both ”False Positive” and ”False Negative” will generate false pseudo labels and reduce performance.}
	\label{fig4}
\end{figure*}

\begin{figure}
	\centering
		\includegraphics[scale=.3]{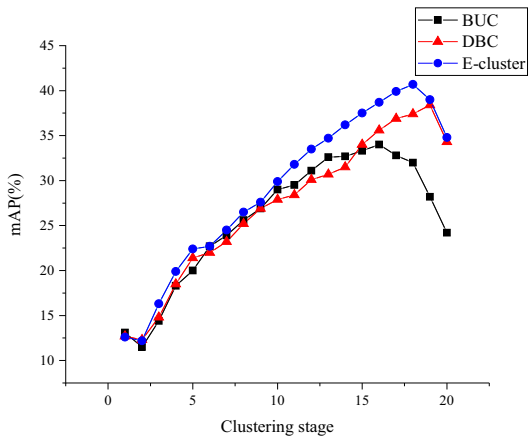}
	\caption{Comparison about results with differnent clustering stages on Market-1501.}
	\label{fig5}
\end{figure}

\section{Ablation Study}
\subsection{Comparison with Hierarchical Clustering Methods}
We compare our method with BUC and DBC on Market-1501 and DukeMTMC-reID in Table \ref{table1}. Both with or without regularization term, E-cluster get the best performance. E-cluster obtains \textbf{rank-1 =70.2\%, mAP =43.0\%} on Market-1501, \textbf{rank-1 =53.2\%, mAP =31.1\%} on DukeMTMC-reID. Compared to BUC, E-cluster gets \textbf{4.7\%} mAP promotion on Market-1501 and \textbf{3.6\%} mAP promotion on DukeMTMC-reID. We can see E-cluster get significant improvement and it proves energy distance does better to performance than minimum distance and UPGMA on re-ID tasks. Besides, it can also illustrate SSD is an appropriate standard to measure the intra-cluster distance because it consider both samples number and the dispersion degree in each cluster. 

\subsection{Comparison with State-of-the-arts}
Table \ref{table2} reports the result of comparison with state-of-the-arts unsupervised re-ID methods. We achieve \textbf{rank-1 = 70.2\%, mAP = 43.0\%} on Market-1501, \textbf{rank-1=53.2\%,  mAP=31.1\%} on DukeMTMC-reID, which surpasses all fully unsupervised methods. Besides, we also compare E-cluster with some UDA methods. Although these methods use source labeled data for pre-training, our E-cluster still performs better than these methods. It illustrates even if without a labeled source dataset, we can generate good pseudo labels through hierarchical clustering with a good distance measurement and get better performance than other methods.

\subsection{Effectiveness of Energy distance}
In order to evaluate the effectiveness of energy distance, we compare E-cluster with BUC and DBC. Figure \ref{fig5} shows the result on Market-1501. We compare training process of three methods by taking mAP as the indicator. We still set the merging percent to 0.05, so there are 20 iterations in total. At first, there isn't much difference among three methods. In the middle stage of training, performance differences gradually increase and E-cluster gets the best. In the end, we observe an obvious performance decline in all methods. But E-cluster still keeps the best performance compared to others.

\begin{figure}
	\centering
		\includegraphics[scale=.3]{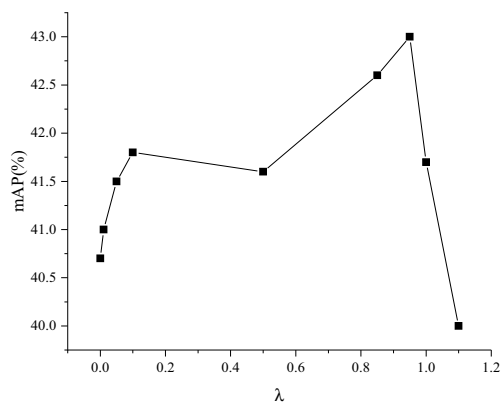}
	\caption{Comparison with different $\lambda$ on Market-1501.}
	\label{fig6}
\end{figure}


It also shows a problem of hierarchical clustering. No matter what distance measurement method is used, the performance of the algorithm will decline significantly at the end of the iteration. We believe that in the later stage of training, there are a lot of big clusters. At this moment, the wrong merging will generate lots of false pseudo labels and lead to a decline in performance. An effective solution is to propose a more effective distance measurement method to reduce wrong merging and finally reduce the decline in later iterations. This is also the next research direction of hierarchical clustering.

\subsection{Trade-off Parameter Analysis}
In order to evaluate the influence of regularization term, we compare the performance with different $\lambda$ on Market-1501. The result is reported in Figure \ref{fig6}. When $\lambda$ increases from 0 to 0.9, the model performance reaches its peak, then further increases $\lambda$ will lead to obvious performance decline. We believe that when $\lambda$ is too small, SSD has little effect in distance measurement. Besides, too large $\lambda$ will lead to high SSD weight in Eq.(\ref{equ:E8}), it will mask the superior properties of the energy distance. So mAP will even be lower than E-cluster without regularization term.

\subsection{Qualitative Analysis of T-SNE Visualization}
As shown in Figure.\ref{fig4}, we compare E-cluster with fully unsupervised cluster-based methods BUC and DBC. We find E-cluster can get more compact clusters because energy distance is a more reasonable distance measurement in re-ID task for hierarchical clustering. However, all these methods have difficulties to distinguish “False Negative” and "False Positive" samples. We believe this will be the research direction of cluster-based methods in the future.

\section{Conclusions}
In this paper, we analyze deficiencies of minimum distance, UPGMA, the number of samples and the dispersion degree used in regularization term. Based on these, we propose E-cluster which combines the energy distance with SSD, we use a trade-off parameter $\lambda$ for balance. The energy distance makes clusters more compact and the SSD considers both the number of samples and the dispersion degree in each cluster, which can further improve model performance. Extensive experiments show that E-cluster has better performance in fully unsupervised methods.

\section*{Acknowledgments}
We thank the reviewers for their feedback. We thank our group members for feedback and the stimulating intellectual environment they provide. This research was supported by The Science and Technology Planning Project of Hunan Province (No.2019RS2027) and National Key Research and Development Program of China (No.2018YFB0204301).

\bibliographystyle{cas-model2-names}

\bibliography{cas-refs}


\end{document}